# McGAN: Generating Manufacturable Designs by Embedding Manufacturing Rules into Conditional Generative Adversarial Network

Zhichao Wang, Xiaoliang Yan, Shreyes Melkote, David Rosen

**Abstract**

*Generative design (GD) methods aim to automatically generate a wide variety of designs that satisfy functional or aesthetic design requirements. However, research to date generally lacks considerations of manufacturability of the generated designs. To this end, we propose a novel GD approach by using deep neural networks to encode design for manufacturing (DFM) rules, thereby modifying part designs to make them manufacturable by a given manufacturing process. Specifically, a three-step approach is proposed: first, an instance segmentation method, Mask R-CNN, is used to decompose a part design into subregions. Second, a conditional generative adversarial neural network (cGAN), Pix2Pix, transforms unmanufacturable decomposed subregions into manufacturable subregions. The transformed subregions of designs are subsequently reintegrated into a unified manufacturable design. These three steps, Mask-RCNN, Pix2Pix, and reintegration, form the basis of the proposed Manufacturable conditional GAN (McGAN) framework. Experimental results show that McGAN can transform existing unmanufacturable designs to generate their corresponding manufacturable counterparts automatically that realize the specified manufacturing rules in an efficient and robust manner. The effectiveness of McGAN is demonstrated through two-dimensional design case studies of an injection molding process.*

## 1. Introduction

In a concurrent design-to-manufacturing pipeline, conceptual design is first created by designers to satisfy various customer requirements. Subsequently, manufacturing engineers are involved in the early stages of design to make suggestions and transform the conceptual designs into manufacturable designs, during which manufacturing processes are selected and design details following certain manufacturability rules are determined. By going through iterations of redesign, a final design is obtained that satisfies both functionality and manufacturability requirements. While this concurrent engineering approach is still prevalent in practice today, the iterative process is time consuming and prone to human errors [1]. It is also noted that the conventional approach may suffer from inadequate expert experience. A novice design engineer may not be able to converge to an optimal design after numerous iterations. Therefore, the inefficiency and the lack of human consistency of the current practice point to the need for an automated approach to create manufacturable designs.

Significant work has been conducted to automate manufacturing process selection [2-5]. After a manufacturing process is selected, the subsequent step is focused on design modification to improve functionality and manufacturability. A plethora of works on GD have focused on design for functionality (DFF), i.e., generating new mechanical designs that have good functional attributes. For example, in the structural design domain, topology optimization has been utilized to minimize the compliance to improve mechanical performance of the structure. In stark contrast, generative design for manufacturability (DFM) has received little attention, as mechanical designs are complex with numerous manufacturing rules to be satisfied. During conventional DFM, engineers generally identify certain features within the design, such as holes, slots, and steps. Then, engineers modify the feature's size, shape, and/or position based on the limitations of manufacturing processes, e.g., rounding sharp corners in injection molded parts. This process of identifying meaningful features, or sub-regions of designs, and modifying them sequentially to satisfy known manufacturability requirements, inspired our proposed automated DFM framework. In this paper, we explore the application of GD methods, specifically Generative Adversarial Networks (GANs), to the problem of modifying part shape details by converting unmanufacturable part designs to manufacturable designs. It has been demonstrated in the literature that GANs are capable of generating a wide variety of diverse images based on a prepared training dataset. We demonstrate that GANs can be trained to embed DFM rules and transform unmanufacturable designs into corresponding manufacturable designs. Specifically, GAN-based image processing methods are utilized to form the proposed DFM workflow: to begin, instance segmentation is utilized to decompose a large complex design into several subregions of interest, from which unmanufacturable subregions are identified; for each decomposed unmanufacturable subregion, a conditional generative adversarial network (cGAN) is then utilized to modify the subregion to ensure its manufacturability; and lastly when all unmanufacturable subregions are modified, a manufacturable part design is obtained by reintegrating all modified subregions.

While under-investigated in the field of DFM, image segmentation is a crucial sub-field of computer vision, where the goal is to identify and combine similar regions within the image that have the same class labels. *Instance segmentation* is a specific sub-class of image segmentation that has three main steps: object detection, object classification, and segmentation. In the context of DFM, object detection refers to identifying the existence of subregions of interest in a complex design, i.e., discovering the existence of a manufacturing feature without explicitly determining the class of the region. Object classification refers to classifying the detected features into a specific class, like slot, blind holes, and steps. Segmentation identifies the shape of the detected object and assigns appropriate class labels to all pixels in the shape. Through the application of instance segmentation, different objects/features in the design and their corresponding class labels are identified, which is important since different segmented regions may require different manufacturing rules. In this work, Mask-RCNN, a well-known instance segmentation algorithm, is utilized to realize the goal (see Section 3.3).

After the original complex design is decomposed into simple regions, subsequent cGANs are utilized to modify and make them manufacturable. In this paper, Pix2Pix, a specific type of cGAN, is utilized to embed DFM rules that modify unmanufacturable part regions. For the purposes of this work, we consider common manufacturing rules associated with injection molding of wall type features in the design. Here, the original unmanufacturable part regions are used as conditional inputs to the GAN. For each type of wall feature, we train a separate Pix2Pix cGAN, which embeds the DFM rule and generates manufacturable shapes based on the unmanufacturable part region. More details of Pix2Pix can be found in Section 3.4.

We name the proposed GAN-based DFM approach "Manufacturable cGAN," or simply McGAN. The structure and application of McGAN are shown in Figure 1. With reference to the McGAN framework in Figure 1a, the Mask R-CNN module segments the design into individual features, given a part design represented as a 2D image, then Pix2Pix modifies each feature to ensure its manufacturability, and lastly, the modified manufacturable features are reintegrated to form the final modified design. While we limit the scope of experimentation in this work to design for injection molding, the proposed framework is general and should be applicable to other manufacturing processes. Specifically, we recognize three types of wall features, thin, thick, and side walls, and three types of design modifications for each wall feature: adjusting feature height and width to satisfy aspect ratio constraints, adding a draft angle, and adding rounds to sharp corners. At the end, the modified features are assembled into a complete part design. Figure 1b shows the application of McGAN to an example part with side walls and one internal wall. This example illustrates the scope and domain of this research. For more complex manufacturing features and processes, a similar scheme can be applied.

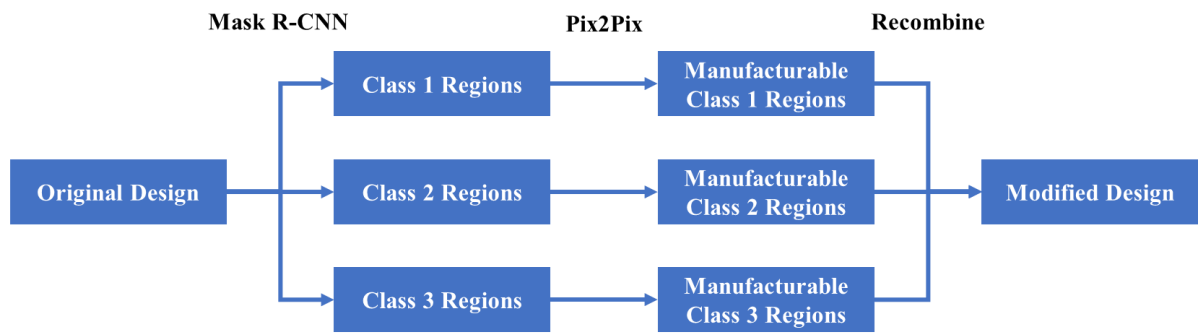

(a). Framework of McGAN

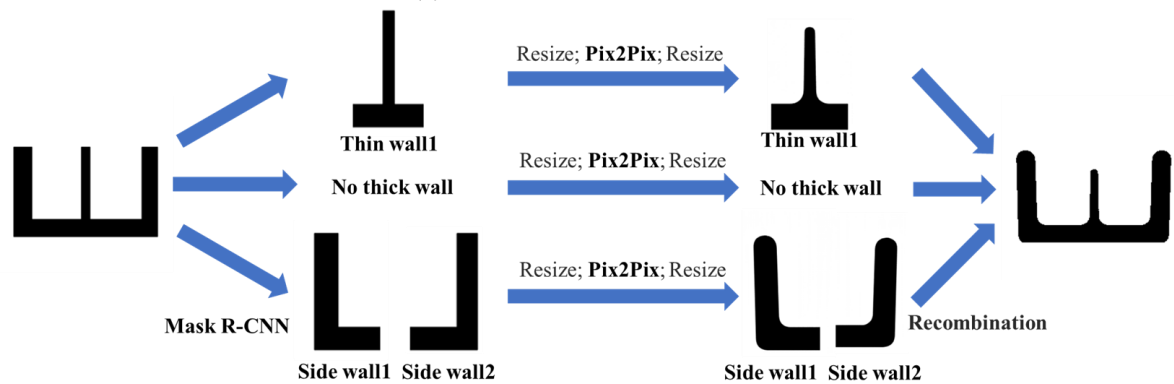

(b). Application of McGAN for Injection molding

Figure 1. Structure and application of McGAN.

To demonstrate the robustness of McGAN at each stage of the framework, several experimental results are reported in the following sections of the paper. The results show that McGAN can robustly modify unmanufacturable designs and generate their corresponding manufacturable counterparts. The rest of the paper is organized as follows. Section 2 reviews research in four different fields relevant to our work: segmentation, GAN, feature recognition and manufacturability analysis, and GD. Section 3 introduces the McGAN framework including the manufacturing rules, the training dataset, Mask-RCNN, and Pix2Pix. Section 4 presents experiments on Mask-RCNN, Pix2Pix, McGAN, and other generative methods. Finally, we will end this paper with discussion and conclusions with future research directions.

## 2. Literature Review

### 2.1. Instance Segmentation

The objective of image segmentation is to group similar regions or segments of an image into classes with respective labels. With the rapid development of deep learning and computer vision, the field of image segmentation has made great improvements in the past decade [6]. Instance segmentation is a subfield of image segmentation, and contains three subtasks: object detection, object classification, and segmentation [7]. A brief review of these three fields is presented in the following.

A major line of research in object detection detects Regions of Interest (ROIs) in images utilizing bounding boxes, and a representative work is the region-based CNN called R-CNN [8]. Compared with R-CNN which detects ROIs from original images, Fast R-CNN detects ROIs from feature maps with RoIPool, and as a result, the inference speed and detection accuracy are improved [9]. Faster R-CNN further speeds up the process of generating ROIs with a Region Proposal Network (RPN) [10]. Object classification identifies the category of a given object. The most classic example of object classification is ImageNet, which identifies the category of an image from 1000 choices like cat and dog [11]. Several benchmarks have been reported in the field of object classification like AlexNet [12] and ResNet [13]. Segmentation labels each pixel of an image with a specific class that it represents. Fully convolution network realizes segmentation through convolutional layers [14]. U-Net creatively utilizes convolution and transpose convolution for segmentation, and the structure of U-Net is broadly utilized in recent neural networks for segmentation [15]. PSPNet proposed the dilated convolution to gain a wide field of view while preserving full spatial dimension [16].

In instance segmentation, the three tasks, i.e., object detection, object classification, and segmentation are conducted together. Mask R-CNN, which combines Faster R-CNN with segmentation, realizes the goal of instance segmentation [17] and is utilized in this paper for instance segmentation. The inputs to Mask R-CNN are complex mechanical designs, while the outputs are the bounding boxes and class labels for each ROI and segmentation of the whole image.

### 2.2. Generative adversarial networks

Generative modeling is an important field of machine learning and has regained prominence in recent years through deep generative learning. Different types of generative models have been proposed such as variational autoencoder (VAE) [18], generative adversarial network (GAN) [19], and diffusion model [20]. VAEs can generate diverse images efficiently, however the quality of generated images may not be good. GANs can generate high-fidelity images efficiently but can be difficult to train, suffering from problems such as mode collapse. Diffusion models can generate diverse high-fidelity images, while the inference speed is generally slow. In this paper, considering the advantages and disadvantages of different generative methods, GAN will be utilized for generative design.

GAN was proposed by Goodfellow et al. in 2014 [19]. A GAN consists of two parts: a generator and a discriminator. The goal of the generator is to create fake data from noise to fool the discriminator, while the discriminator distinguishes between real data collected in advance and fake data generated by the generator. During the training process, the generator and discriminator compete and improve their performance simultaneously. After the training process, the generator can generate realistic new images from noise. Since its introduction, GAN has greatly improved from the viewpoints of quality and diversity. Some benchmarks works have been established like EBGAN [21], BEGAN [22], Self-Attention GAN [23], and BigGAN [24].

All of the previously mentioned works utilized unconditional GANs, where image generation is not conditioned by any inputs to the generator. In contrast, conditional GANs (cGAN) are controlled by the input condition. Of interest here is the use of an input image as the condition, where that image is modified by the cGAN. Mirza and Osindero [25] applied a conditional GAN on MNIST with digits as the condition. In comparison, Pix2Pix [26] and CycleGAN [27] utilize images as the condition to achieve image-to-image translation. Pix2Pix is a supervised neural network for image-to-image translation, while CycleGAN is an unsupervised network. In this paper, Pix2Pix is applied to translate unmanufacturable designs into manufacturable designs.

As mentioned, GANs can be difficult to train, with vanishing gradient and mode collapse as two common problems. Vanishing gradient refers to the gradient of the loss function going to zero, meaning that the weights and biases of the loss function cannot be modified. Mode collapse means that generated designs represent only a subset of the original training subset, i.e., the diversity of the training dataset is lost. To deal with these two problems in GAN, different loss functions have been utilized to stabilize the training process. WGAN-GP [28] and LSGAN [29] incorporate frequently utilized loss functions to solve these problems. In this paper, the loss function from LSGAN is utilized.

### 2.3. Feature recognition and manufacturability analysis

Feature identification for manufacturing research started in the 1980s when researchers started investigating automation of the design-to-manufacturing workflow through manufacturing features recognized automatically from design models. Different types of methods have been investigated including graph-based, hint-based, and volumetric-based methods [30, 31]. The graph-based method transforms CAD models from a boundary representation to a graph representation, and then graph feature extraction techniques are utilized. Hint-based methods are designed to deal with intersecting features. Volumetric-based methods decompose a solid model into simpler shapes for feature extraction. Different from previous work on global shape description, some research has been conducted to extract local features. Curvature is a simple but useful tool for local feature extraction. D2 utilizes shape distributions to describe 3D models [32]. KNN was utilized to cluster features using an unsupervised learning approach [33]. The support vector machine method was utilized to extract features by Ip et al. [34]. Heat kernel signature was utilized as a shape descriptor to support redesign and subsequent process planning [35]. The artificial bee colony algorithm and back propagation neural network were combined to extract geometric information like "step", "boss" and "slot". A 3D CNN was utilized to extract features from voxelated CAD models that were applied for classifying manufacturability [36]. These works could identify features of CAD models but could not directly modify models to improve their manufacturability.

Design for manufacturing refers to designing parts that satisfy manufacturability requirements of a given process [37]. The DFM literature is extensive, but we are interested in computational approaches for modifying geometric part models to meet manufacturability requirements. For casting, a virtual thermal diffusion problem was defined, and a global thermal constraint was added into the optimization formulation to guarantee the generated model satisfies cast-ability requirements of no undercuts or cavities along a prescribed draw/mill direction [38]. For simple machining, i.e. 2.5D profiling, a feature-based shape optimization method was proposed by incorporating a feature-fitting algorithm into a level-set topology optimization (TO) method to generate manufacturable designs [39]. For multi-axis machining, an inaccessibility measure field was defined over the design domain to identify non-manufacturable features, and it was utilized to penalize the sensitivity field so that the outputs of TO are manufacturable [40]. Another work added a multi-axis constraint for enforcing manufacturability of a TO design [41]. For additive manufacturing (AM), a surrogate model based on neural networks was utilized to predict maximum shear strain index to ensure manufacturability with respect to cracks [42]. A systematic approach was developed to ensure the removability of support structures of TO for AM [43].

## 2.4. Generative design (GD)

Pioneering research has been conducted on GD that combines deep neural networks with TO to generate a wide variety of designs that satisfy performance requirements. Previous researchers utilized iterations between GAN and TO to generate functional designs [44]. Another work utilized reinforcement learning (RL) with maximization of the diversity of topology designs as reward functions [45]. Yamasaki et al. [46] applied VAE to generate new designs with a sensitivity-free and multi-objective method. Physical fields can also be added to the generator of a cGAN to accelerate topology optimization. Our previous work embeds topology optimization into GAN to avoid iterations between GAN and TO and accelerate the generation process of new designs [47]. All these papers work on DFF in the 2D domain, while a small number of papers focus on 3D generative DFF. 3D models can be generated using GAN and evaluated under different complex simulations [48]. A general CAD/CAE framework for automatic 3D CAD designs and performance evaluation was put forward by Yoo et al. [49].

Research to date on generative DFM is scarce. Williams, Meisel, Simpson and McComb [50] trained a 3D CNN to evaluate the manufacturability of generated additive manufacturing designs. Guo, Lu and Fuh [51] trained an autoencoder-generative adversarial network to evaluate the manufacturability of metal cellular structures in the metal powder bed fusion process. Another direction utilized manufacturable designs to train GAN. Manufacturing constraints of a 3-axis computer numerically controlled (CNC) milling machine were taken into consideration when collecting part models for the training dataset [52]. The trained GAN could generate manufacturable designs. However, no paper focused on utilizing GD to modify the unmanufacturable regions of designs to make them manufacturable, which is the main focus of this paper.

## 3. McGAN

In this section, the details of our generative design McGAN approach are provided. We start with an introduction on the general structure of McGAN that transforms unmanufacturable designs into manufacturable designs. After that, the injection-molding dataset for training and testing McGAN will be introduced. In addition, a successful application of instance segmentation utilizing Mask R-CNN will be demonstrated. Finally, we delve into the Pix2Pix cGAN model, which utilizes images as its conditional input.

### 3.1. Method Overview

McGAN can modify existing unmanufacturable designs and generate their corresponding manufacturable counterparts. The structure of McGAN was presented and illustrated in Figure 1. It is composed of three parts: Mask R-CNN, Pix2pix and reintegration. To begin with, the input part design is sent to Mask R-CNN for instance segmentation. The output of Mask R-CNN includes regions that each include one potentially unmanufacturable thin wall, thick wall, or side wall. For the example shown in Figure 1, Mask R-CNN successfully identified one thin wall and two side walls named "thin wall 1", "side wall 1" and "side wall 2", respectively. Since the x-axis and y-axis limits of the outputs of Mask R-CNN are different from those of the inputs to the Pix2Pix cGANs, a translation and resize scheme was utilized to translate and scale the segmented walls to produce the output of Mask R-CNN. Then, these segmented features are sent to Pix2Pix for modification (note that we will use "Pix2Pix" instead of "the Pix2Pix cGAN" for brevity). Note that three different Pix2Pix models, for the three different features, were trained to transform unmanufacturable designs to manufacturable designs. After modifications, the manufacturable features are translated and rescaled to the original size. The last step of McGAN recombines the modified manufacturable wall features to form the final design.

### 3.2. Injection Molding Datasets

In this subsection, the two injection molding datasets that were used to train and test Mask R-CNN and Pix2Pix modules, respectively, are described. Three types of injection molding features were considered: thin wall, thick wall, and side wall. The domain of part models was limited to simple rectangular-shaped product housings or variants of housings. Each part design has a bottom wall that is held constant, while other features extend above the bottom wall. Side walls are vertical walls at the ends of the bottom wall and represent the sides of a housing. The thin and thick features are internal features, i.e., they lie in the middle regions of the bottom wall. These internal features could model internal walls, ribs, bosses, and similar protrusions.

Three types of injection molding rules were implemented, specifically adding draft angle to facilitate part removal from the mold, rounding corners and edges to aid mold filling and part removal, and vertical feature aspect ratio limits to eliminate feature breakage during ejection. These rules are very common in DFM handbooks and rule sets [53].

The first dataset was utilized to train and test the Mask R-CNN instance segmentation module. As shown in Figure 2, complex wall examples were composed of three of the simple walls mentioned before (thin, thick, or side wall). The "Image" in the left column represents the part model and is the input to Mask R-CNN, while the "Mask" is the ground truth used to train Mask R-CNN. The dimension of the "Mask" is the same as that of the "Image". For the background pixels in "Image", their corresponding pixel values in "Mask" would be 0. A simple coding scheme for masks was developed to assist instance segmentation. Each wall type was represented by a two-digit code, where the first digit represents wall type (1, 2, or 3), while the second digit numbers the walls from 1 to 9. These two-digit codes then become the pixel value assigned to pixels that comprise the feature; that is, the codes represent classes. They are also used as pixel intensity in mask images. For thin walls, for example, the pixel values ranged from 11 to 19, where 11, 12 and 19 represented thin wall 1, thin wall 2 and thin wall 9, respectively. The same rules apply for thick walls (21 to 29) and side walls (31 to 39). Since these pixel values represent dark intensities, the third column in Figure 2 labeled "Mask: for visualization" was included so that the mask could be better visualized. Both the "Image" and "Mask" are represented as $256 \times 256$ images. However, for the purposes of specifying manufacturing rules, part images and masks were scaled so that they were 10x10 units. The x-axis and y-axis limits were selected to be [-5, 5] and [-2,8], respectively. As a result, 256 pixels represented 10 times the unit length, i.e., the scale was $\frac{10 units}{256 pixels}$.

For training Mask R-CNN, 5,000 pairs of "Image" and "Mask" were generated. After the training process, 500 new complex "Images" were utilized to test McGAN. These new 500 examples came from the same distribution as the previous 5,000 examples used for training to ensure a robust evaluation of our model's performance.

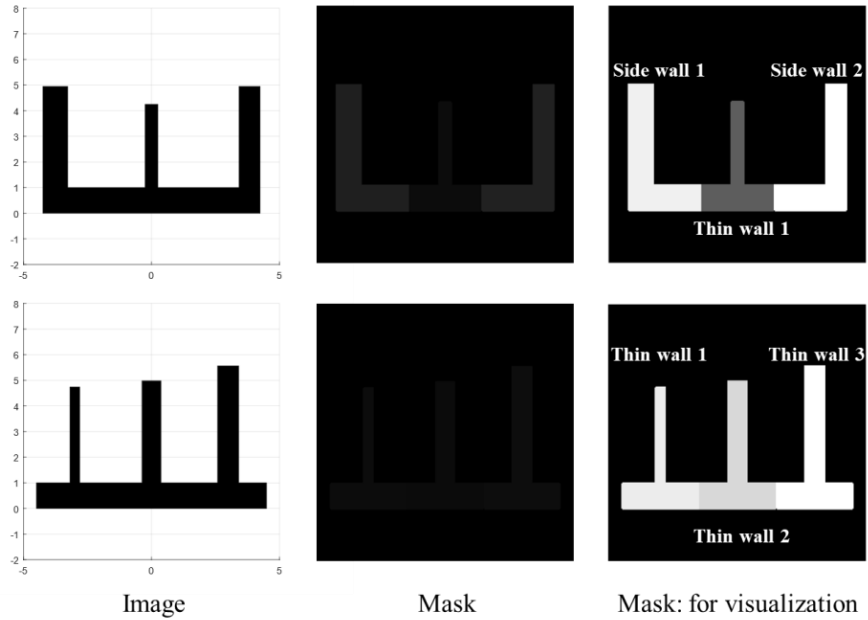

Figure 2. Image and corresponding Mask for training and testing Mask R-CNN.

The second dataset was utilized to train and test manufacturability design modifications using Pix2Pix. For different types of walls, the specific rules for manufacturing were also different. To facilitate the modification process, feature images were scaled up so that they were represented by more pixels than in the original part images, specifically they were 6.6 units in horizontal and vertical directions, i.e., the scale was $\frac{6.6 units}{256 pixels}$. At this scale, bottom walls were always set to be 1 unit thick.

As mentioned, three types of manufacturing rules were considered. These rule types are always applied sequentially for each identified feature: wall aspect ratio, adding draft, and adding rounds to corners. Figure 3 shows examples of each wall type and the modification sequence that they undergo. For thin walls, the aspect ratio rule specifies that the height should be less than 4 times the bottom wall thickness and the vertical wall width should be between 0.4 and 0.6 relative to the bottom wall thickness. For thick walls, the modified height should also be less than 4 and an inner hole is added underneath the wall (appears as a slot in the 2D part images), which limits the width of each thick wall to ensure a uniform cooling process in injection molding. After adding the inner hole, updated wall widths should be between 0.4 and 0.6. For side walls, their heights are not modified since we assume that part housing sizes cannot be changed, while their widths are modified to unit thickness. For the draft angle rule, draft was set as $1°$ for thin and thick walls, while it was set as $1.5°$ for side walls. Lastly, the round corner rule changes sharp corners to rounded ones, with the radius modeled as a random variable between 0.4 to 0.6. For training, 4,000 examples of each wall type were provided to each of three separate Pix2Pix networks, one for each wall type.

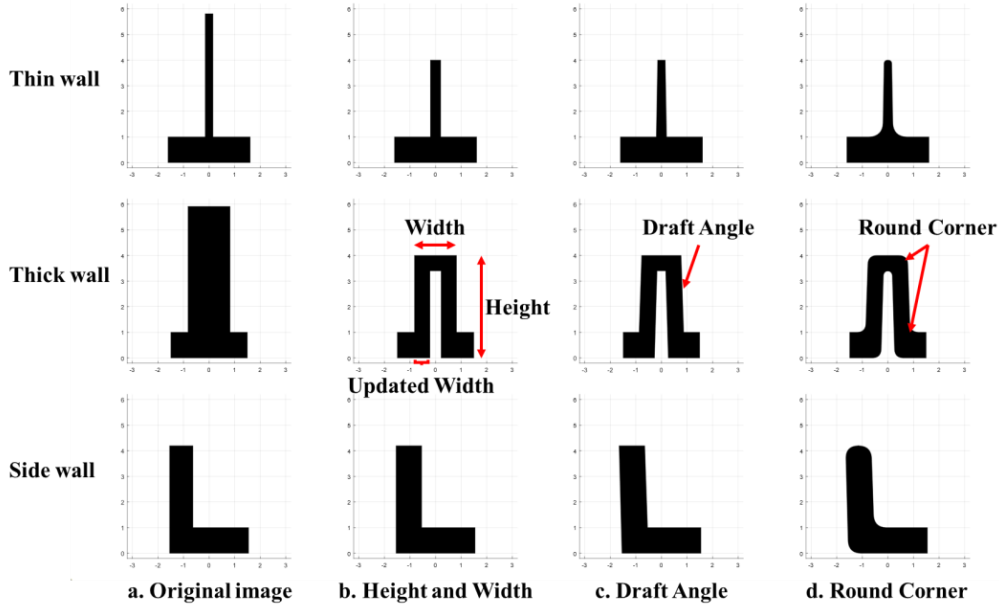

Figure 3. Injection-molding dataset. (a) Original image that is unmanufacturable and is input to Pix2pix. (b) Modify aspect ratio from part (a). (c) Add draft angle to part (b), (d) Add round corner to part generated in (c) and will be utilized as the output of the sequence of modification steps.

### 3.3. Mask R-CNN

In this subsection, the details of instance segmentation achieved by Mask R-CNN are illustrated. The structure of Mask R-CNN is shown in Figure 4. Given a part image as input, a CNN called ResNet-50-FPN is used as the backbone for feature extraction [54]. This CNN is known as a fast, effective method and was used in the original development of Mask R-CNN to produce feature maps across a range of size scales. Note that the term "feature map" has a specific meaning in image processing: it denotes the output of a layer in a CNN and contains regions that indicate activation of certain regions of an image.

These feature maps are then sent to both a ROI proposal network and to a module that extracts features in these ROIs. During training, the ground truth is utilized as an input to the ROI proposal network so that MASK R-CNN can distinguish between good and bad candidate ROIs. After extracting promising regions from the image, the object classification module assigns a probability score to each ROI, indicating the likelihood that it contains a manufacturing feature. A probability score is computed using a sigmoid function at the end of the neural network, ensuring that the resulting probability falls within the range of [0, 1]. Note that all this processing takes place using feature maps, not the original part image. In the last step, the classified ROI feature maps and probability scores are mapped to images, then assembled into the part image in the form of a mask. Several of the steps will be covered in more detail.

For testing, ROIs and corresponding probability scores of a given test image then need to be found. Then, the ROIs in the test image are combined to generate the mask, which constitutes the output of MASK R-CNN.

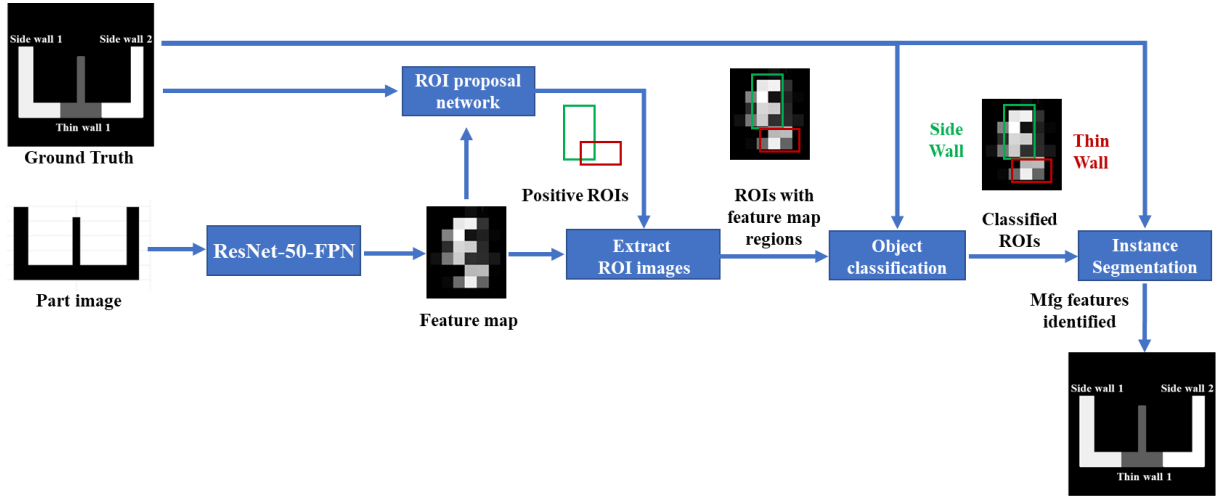

Figure 4. Architecture of Mask R-CNN for this work.

Given the image features extracted from ResNet-50-FPN as shown in Figure 4, the region proposal network is utilized to generate candidate ROIs using Faster R-CNN [10]. Suppose that the feature map is a 2D matrix $f \in \mathbb{R}^{m \times n}$ where $m$ and $n$ are the height and width of the feature map, respectively. For each pixel in the feature map, 9 different candidate ROIs are generated using all combinations of three scales and three ratios as shown in 5(a). In total, $9 \times m \times n$ possible candidate ROIs as shown in the bounding boxes are generated. Then, filtering techniques are applied to delete some ROIs that lie beyond the boundary of the feature map. For each candidate ROI that remains after filtering, an intersection over union (IOU) operation is calculated with respect to the ground truth. The IOU equation is shown in Eq. (1), where $Area(\cdot)$ calculates the area of the region and $R_1$ and $R_2$ are two candidate regions. Because $Area(R_1 \cap R_2) \leq Area(R_1 \cup R_2)$, the resulting $IOU \in [0, 1]$. During training, one of $R_1$ or $R_2$ is a generated ROI, while the other is from the ground truth. As a result, Mask R-CNN learns how to find manufacturing features in typical part images.

$$IOU = \frac{Area(R_1 \cap R_2)}{Area(R_1 \cup R_2)} \quad (1)$$

A candidate ROI is classified as positive when $IOU \geq 0.5$, but negative when $IOU \leq 0.5$ as shown in 5(b). From all candidate ROIs, 128 positive and 128 negative ROIs were selected to train the ROI proposal network.

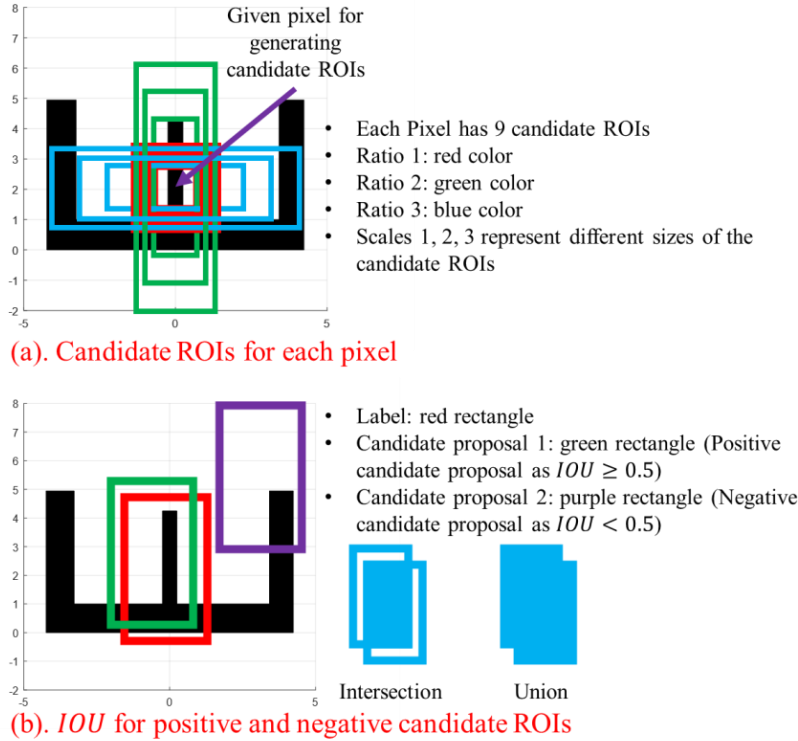

Figure 5. (a). candidate ROIs generation; (b). $IOU$ to classify positive and negative candidate ROIs

The next step applies the ROI, which is represented as a rectangle bounding box, to the feature map, which extracts the region in the feature map that corresponds to the ROI. Some pre-processing must be conducted to scale and translate these regions since subsequent steps require inputs to be of the same size, and this process is termed as ROI align in Mask RCNN. These extracted regions are fed into the object classifier that recognizes regions as representing a thin, thick, or side wall. This is performed using a standard classification CNN.

Finally, the classified feature maps are sent to the instance segmentation module that segments the input image. That is, instance segmentation essentially recognizes each feature as distinct from other features of the same class. For example, the part in Figure 2 (top) and Figure 4 has two side walls, labeled as Side Wall 1 and 2, and one thin wall. In comparison, traditional semantic segmentation can only identify side walls or thin walls regions. As a final step, the instance segmentation module maps each feature into an image (i.e., reverses the actions of ResNet-50-FPN), assembles the images into the entire part image, and applies the segmentation results to produce the final mask.

### 3.4. Pix2Pix

Pix2Pix is a type of cGAN used in this work to modify unmanufacturable designs using the rules presented in Section 3.2. The Pix2Pix framework and structure of the generator and discriminator are shown in Figure 6. The generator generates designs, while the discriminator discriminates between a real design and a fake design generated by the generator. During the training process of the GAN, the generator and discriminator of GAN compete, and as a result, the capabilities of both improve. After training, the discriminator is deleted for Pix2Pix, and the generator alone is utilized to generate new designs.

Different from vanilla GANs which generate new designs from noise, Pix2Pix is a type of conditional GAN that modifies an input image. In this way, it realizes the goal of image-to-image translation and, specifically for this work, modification of unmanufacturable regions of part designs. The details of the Pix2Pix generator can be found in Figure 6b. The general structure of the generator utilizes the "U-Net" structure [15]. As the character "U" suggests, the architecture consists of a left side (convolution layers) and a right side (transpose convolution layers). In the convolution layers, the dimensions of the feature maps decrease, while in the transpose convolution layers, the dimensions of the features increase. In addition, the extracted features in convolution layers are concatenated and utilized in the transpose convolution layers for image detail modification, and this is realized through a copy process which is named the "skip connection." These skip connections can significantly facilitate training in large networks, including partially alleviating the vanishing gradient problem. For all layers in the generator, the convolution parameters used have a kernel size of (4, 4), a stride size of (2, 2) and a padding size of (1, 1).

The discriminator of Pix2Pix can be found in Figure 6c. In the first three convolution layers, the same convolution parameters are used as in the generator. In the last two convolution layers, the stride size is changed to (1, 1). To achieve the objective of image-to-image translation, the discriminator evaluates whether a partial region, called a patch, is real or fake, rather than evaluating the entire input image. This is called the "PatchGAN" style of cGAN. In our work, we divide part images into $30 \times 30$ patches. Hence, the discriminator produces an output of dimension $30 \times 30 \times 1$, and each value in the output is a Boolean value representing real (true) or fake (false). It has been shown that the "PatchGAN" style can improve the performance of GAN in generating realistic designs [26].

To embed Pix2Pix in our McGAN framework, the loss function of the least-squares GAN (LSGAN) was applied for training. For the discriminator of Pix2Pix, the mean square error (MSE) loss was utilized as shown in Eq. (2), where $n_1$ and $n_2$ were the numbers of real and fake images, respectively, $D(\cdot)$ and $G(\cdot)$ represent the generator and the discriminator respectively, $z$ was the input to the generator (unmanufacturable designs) and $x$ was the real data (manufacturable designs). During the training of the discriminator, the generator $G(\cdot)$ was fixed, since only the discriminator $D(\cdot)$ was updated.

$$L_D = \sum_{i=1}^{n_1} \|D(x) - 1\|_2^2 + \sum_{j=1}^{n_2} \|D(G(z)) - 0\|_2^2 \qquad (2)$$

For the Pix2Pix generator, an $L_1$ penalty term was added to the loss function to improve the stability of the cGAN, as seen in Eq. (3).

$$L_G = \sum_{j=1}^{n_2} \left( \|D(G(z)) - 1\|_2^2 + \lambda \times \|G(x) - x\|_1^1 \right) \quad (3)$$

Since $x$ represents the manufacturable image, the generator should not change it. As a result, $G(x)$ is supposed to be equal to $x$, and the $L_1$ penalty term is added to realize this goal. $\lambda$ is a weighting term to balance the two terms within the generator loss function. During the generator training process, the discriminator $D(\cdot)$ was fixed, since only the generator $G(\cdot)$ was updated.

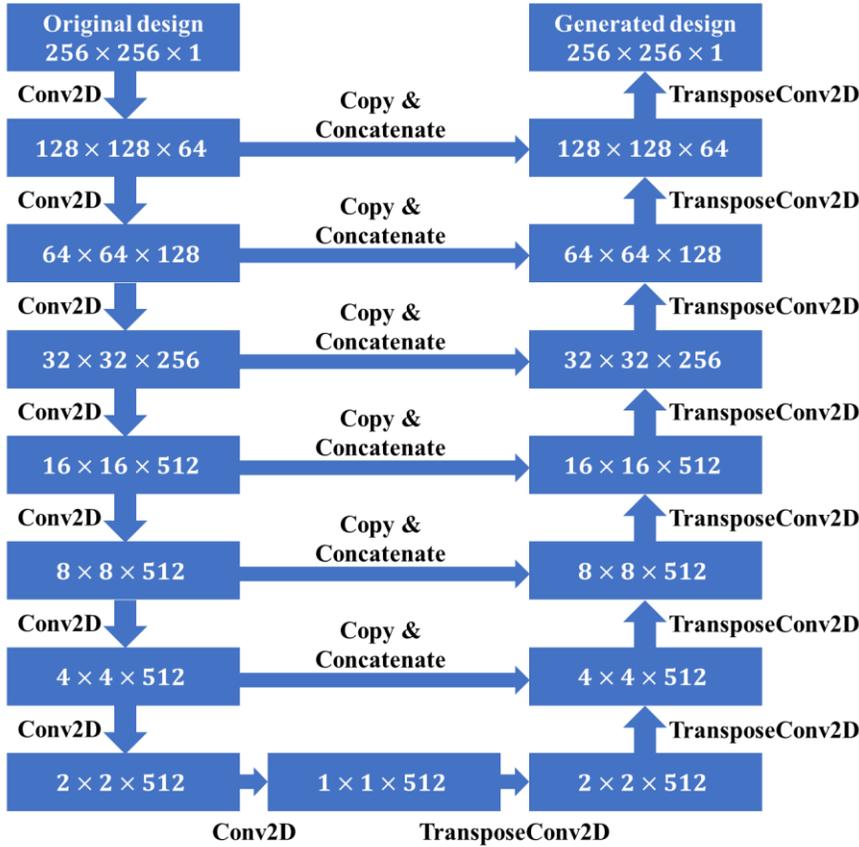
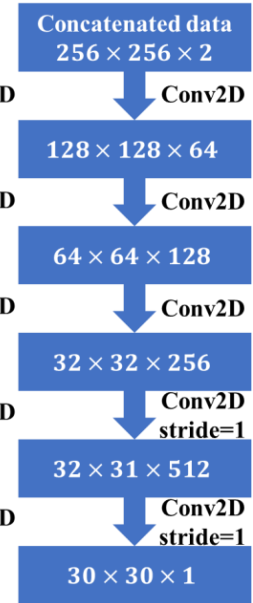

Figure 6. Part (a): the framework of Pix2Pix, including part (b) the generator and part (c) the discriminator

## 4. Experiments

In this section, different experiments were performed to test the performance of Mask R-CNN, Pix2Pix, and McGAN in modifying the unmanufacturable sub-regions of original designs. In addition, comparisons with other methods on modifying unmanufacturable regions are also provided. Lastly, we conduct experiments on different scales of unit thickness, segmentation using short bottom walls, and more complex test data with five sub-walls. Results show that McGAN can modify unmanufacturable designs robustly and generate their corresponding manufacturable counterparts efficiently.

### 4.1. Results of Mask R-CNN

The capabilities of Mask R-CNN to segment part designs with three vertical walls into individual feature sets were tested using 500 example designs each consisting of three vertical walls, for a total of 1500 walls.

Quantitative evaluation of Mask R-CNN, i.e., object detection and semantic segmentation, is provided in Table 1. Let us start with the explanation of $AP_{50}$, which refers to the mean average precision on $IOU$ with a threshold of 50%. For each design, Mask R-CNN predicts some bounding boxes for ROIs. In the meantime, the correct bounding boxes for ROIs are gathered in advance. For each predicted bounding box, its $IOU$ with the correct bounding boxes is computed as shown in Eq. (1). If the computed $IOU$ is larger than the threshold, i.e., 50%, it will be identified as true positive ($TP$). Otherwise, the identified bounding box will be considered as false positive ($FP$). Eventually, the precision for this design is computed as precision $= \frac{TP}{(TP+FP)}$. Lastly, $AP_{50}$ is computed as the average of precision across different designs. The difference between $AP_{50}$ and $AP_{75}$ is the threshold for determining true positive ($TP$) and false positive ($FP$). The metric AP or mAP is an average of multiple $AP_x$ where $x = 50, 55, 60, \cdots, 95$. Lastly, $AP_{small}$, $AP_{medium}$ and $AP_{large}$ are mean precision for small, medium, and large features. Following the COCO dataset, areas are termed 1) small, 2) medium and 3) large if the number of pixels is 1) smaller than $32^2$, 2) larger than $32^2$ but smaller than $96^2$, and 3) larger than $96^2$ [17]. Most of the evaluation metrics, i.e., $AP_x$ are high, and this means that Mask RCNN accurately detects and segments different types of walls. In addition, the results of object detection are better than semantic segmentation because they have higher values for the evaluation metrics and will be utilized for the next tasks for design modification. $AP_{small}$ is absent because there are no small features in our designs.

Table 1. Quantitative evaluation of Mask R-CNN

|  | AP (mAP) | $AP_{50}$ | $AP_{75}$ | $AP_{small}$ | $AP_{medium}$ | $AP_{large}$ |
|---|---|---|---|---|---|---|
| Object detection | 99.9 | 100 | 100 | - | 100 | 99.9 |
| Semantic segmentation | 97.3 | 100 | 100 | - | 96.1 | 98.4 |

During these tests, some errors occurred in the output of Mask R-CNN as shown in Figure 5. The first problem is named "multiple identifications" by the authors. As shown in the figure, the right side of the complex design was identified as both a "thin wall" and a "thick wall" at the same time. This was caused by the width of the right-side sub-wall being close to the thickness threshold between thin and thick walls. As a result, it was difficult for the right-side sub-wall to be accurately classified. The second problem occurred when Mask R-CNN tried to cut the internal thin wall into two parts and identified the left side of the thin wall as another "side wall". Though the frequency of these two problems was low (57 mistakes in the 1500 simple wall examples), some filtering methods were developed to delete these regions that were mislabeled or misidentified.

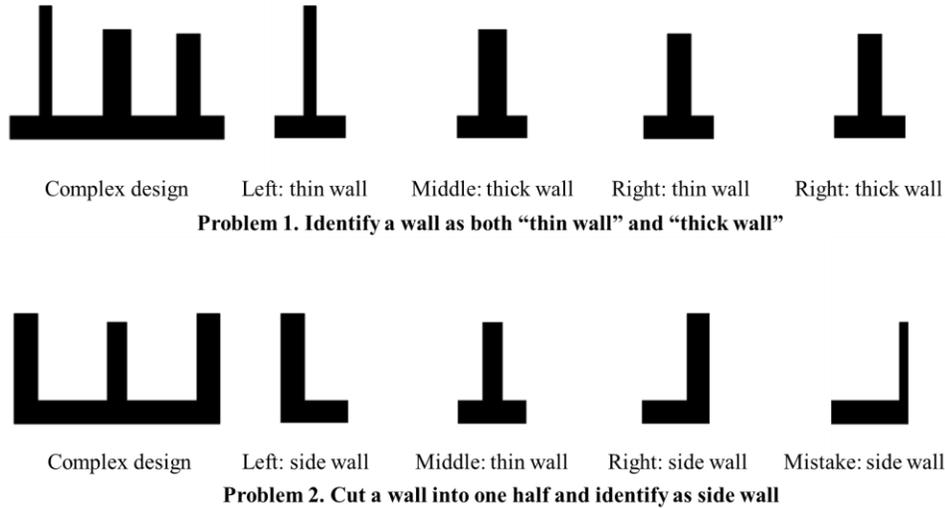

Figure 5. Two problems for Mask R-CNN in segmentation complex designs with three sub-walls

We utilized the $IOU$ index in Eq. (1) to avoid repeated identification of a wall as both "thin wall" and "thick wall," presented as problem 1 in Figure 5, and the threshold of $IOU$ is set as 0.2. For a complex design, Mask R-CNN firstly segmented the original design into multiple sub-designs, with a probability score computed for each sub-design. The probability score measures the probability of the identified bounding box being a true ROI. The smaller the probability score, the less likely the bounding box will be a ROI. Suppose that there are $n$ bounding boxes from the output of Mask R-CNN. The bounding boxes and their corresponding probability scores can be represented as $BB_i$ and $p_i$ for $i = 1,2,\cdots,n$. For each pair of bounding boxes like $BB_i$ and $BB_j$, the value of $IOU_{ij}$ is calculated according to Eq. (1). If the calculated $IOU_{ij}$ is larger than the threshold of 0.2, the bounding box with a smaller probability score $p = min(p_i, p_j)$ will be deleted. Note that there are $\frac{n \times (n-1)}{2}$ pairs of bounding boxes in total. Using this filtering technique to delete repeating bounding boxes, all 1,500 sub-designs were correctly identified. Some results of the segmentation and object detection can be found in Figure 6. As shown in the image, the segmentation results may be partially blurry at the boundary as shown in the red rectangles, and this is natural for the segmentation problem and it is also consistent with the quantitative evaluation in Table 1. But the results of segmentation and object detection were of high quality and could be utilized in McGAN for modification tasks by Pix2Pix. To facilitate recombination of manufacturable features derived from Pix2Pix, the bounding boxes for object detection were extended by 5 pixels in the four directions, i.e., top, bottom, left, and right. In this way, different bounding boxes will have some overlap, and as a result, no gaps will form during the recombination process in McGAN as shown in section 4.3.

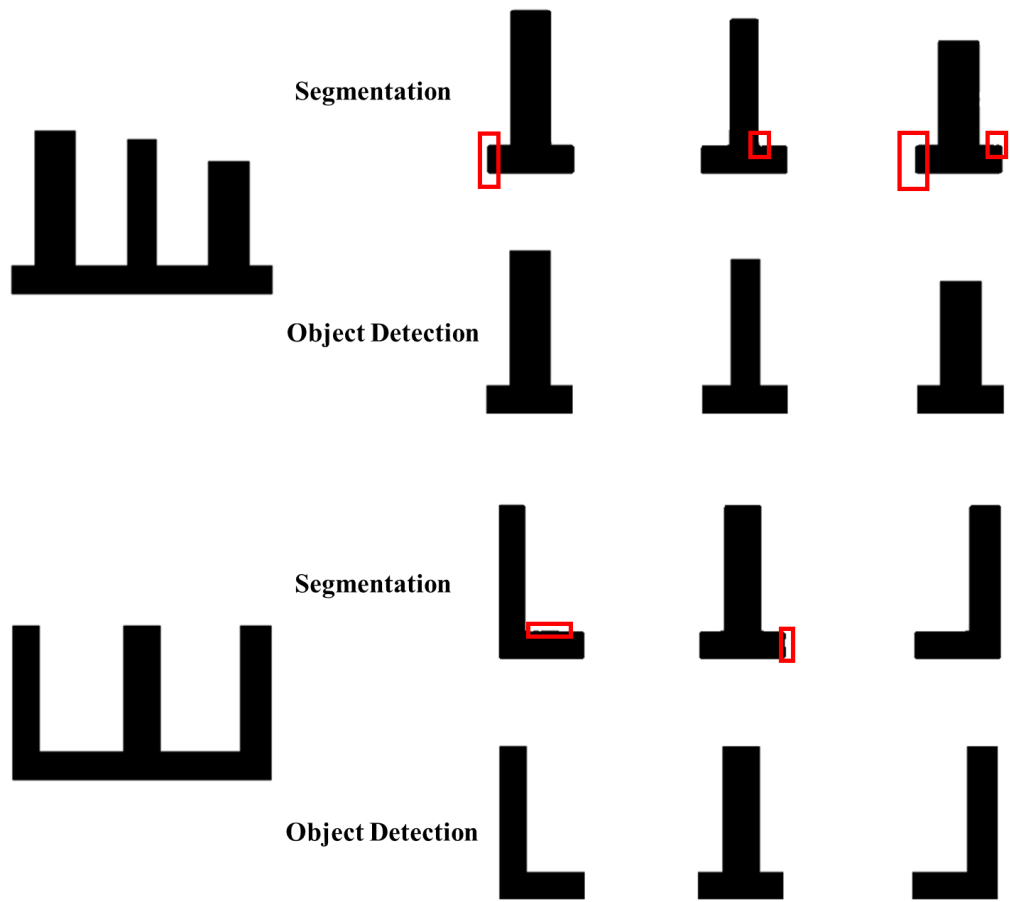

Figure 6. Segmentation and object detection of the original design

## 4.2. Results of Pix2Pix

In this subsection, the results of Pix2Pix are presented. Three Pix2Pixs were trained separately to modify thin, thick, and side walls, respectively. The results of the modified walls can be found in Figure 7. As the figure shows, the unmanufacturable regions of the original designs were replaced with manufacturable designs. For the thin walls, the height and width of the upper walls were modified. Draft angles and round corners were also added to facilitate part removal from the mold. As for thick walls, the heights of the upper walls were modified using the same scheme, while an internal hole was added to modify the width. The processes of adding draft angles and round corners for thick walls were similar to those of thin walls except that they were also applied to internal holes. Different from thin walls and thick walls, side walls do not require modifications of height, while the side wall widths were modified. The process of applying draft angles was also different as the draft angles are added in the outward direction while the draft angles for thin and thick walls are in the inward direction. The process of adding round corners for side walls was the same as that of thin walls or thick walls. The results showed that the modification process of Pix2Pix was robust and efficient.

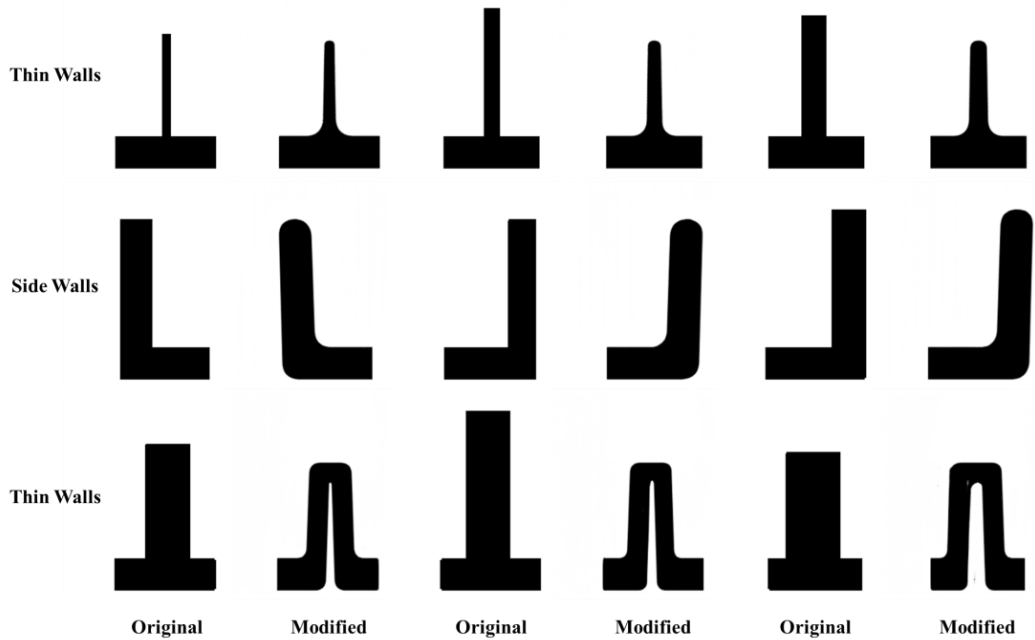

Figure 7. Modified unmanufacturable regions using three separate Pix2Pixs

Evaluating Pix2Pix poses a more intricate challenge compared to Mask R-CNN. The original Pix2Pix paper introduces two evaluation metrics: 1. human evaluation via Amazon Mechanical Turk and 2. Semantic segmentation score [26]. The first method involved crowdsourcing evaluation, but assessing modified designs in this work demands manufacturing expertise that crowd members may not have. The second approach employed a separate fully convolutional network (FCN) for semantic segmentation. However, our modified design, being a binary image with a single feature, makes segmentation relatively straightforward. Consequently, the second method is not suitable for our task. In addition, common metrics in the spatial domain like Peak Signal-to-Noise Ratio (PSNR) and Structural Similarity Index (SSIM) might not be appropriate either. Unmanufacturable designs offer numerous choices for modification, such as various draft angles and round corner radii. The provided label is a suggested solution rather than the sole solution. Lastly, a feature-level evaluation metric like Frechet Inception Distance (FID) is more suitable. FID assumes the feature space resembles a high-dimensional Gaussian distribution. It extracts features from hundreds of images in two sets—true images and generated images—then calculates distances using the Earth mover's distance measure. If two sets of images are similar, the computed FID should be smaller. The FID results can be found in Table 2. For the test dataset, we have "input" which is sent to Pix2Pix for modification and corresponding "label", which is a modified version of "input" to make it manufacturable. The "output" refers to the designs generated from the "input" after modification by Pix2Pix. We have computed FID between "input" and "label" and FID between "output" and "label". Because Pix2Pix has been trained to modify the unmanufacturable designs and make them manufacturable, the FID score between "output" and "label" is much smaller than the FID score between "input" and "label".

Table 2. Quantitative evaluation of Pix2Pix

| Evaluation metric | Thin walls | Thick walls | Side walls |
|---|---|---|---|
| FID between output and label | 29.72 | 13.79 | 15.60 |
| FID between input and label | 205.62 | 188.96 | 222.90 |

The previous process of modification utilized three Pix2Pixs to modify the three wall types. However, it is also possible to train one Pix2Pix to modify all three wall types together. The results of applying one Pix2Pix to modify different types of walls can be found in Figure 8, Although the results are mostly correct, it seems that Pix2Pix occasionally confuses between thin walls and thick walls. For example, Pix2Pix tries to add internal holes to thin walls, and this is consistent with the authors' observation that when wall widths lie near the threshold between thin and thick walls Pix2Pix can get confused. Also because of the training of all three types of walls together, the modification of thick walls deteriorated as shown in the last example of thick walls. Although training of the three wall types together was mostly successful, the previous scheme of training three separate Pix2Pixs was demonstrated to be more robust and will be utilized going forward.

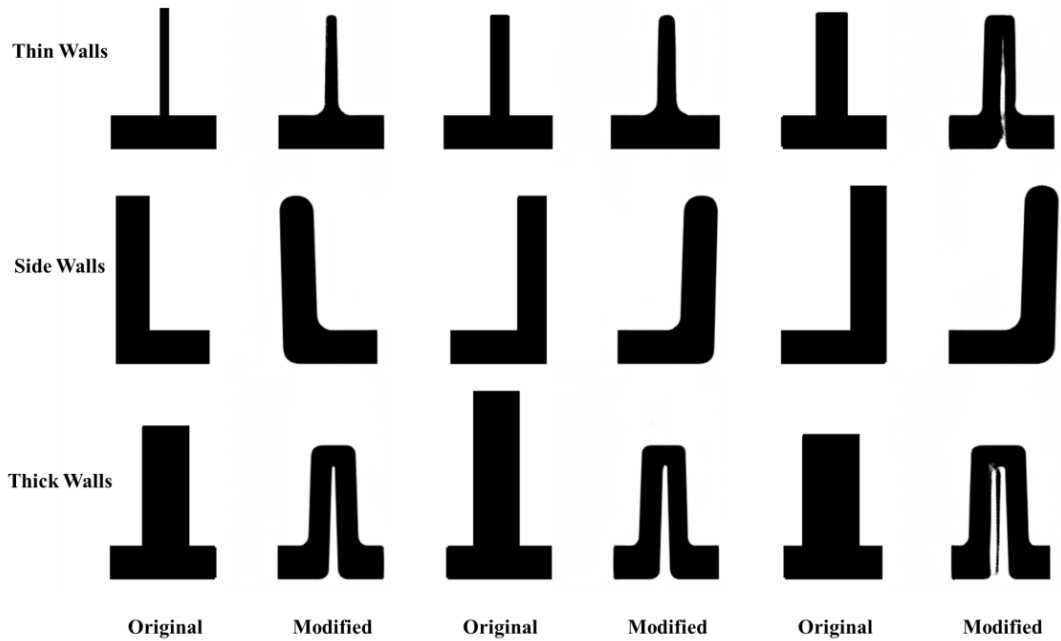

Figure 8. Modified unmanufacturable regions using one Pix2Pix

## 4.3. Results of McGAN

As shown in part (b) of Figure 1, the example composed of three sub-walls was firstly segmented into three regions through Mask R-CNN, and each segmented sub-wall was classified as thin wall, thick wall, or side wall. A resizing scheme was performed to ensure the segmented data were consistent with the training dataset. Pix2Pix was then utilized to modify the unmanufacturable regions and generate new manufacturable regions. The last step was the resizing and recombining to form the final complex designs that are manufacturable.

More examples generated by McGAN can be found in Figure 9. The red bounding boxes refer to the bounding boxes of the original unmanufacturable designs. As shown in the top three examples of Figure 9, all these walls were either thin walls or thick walls, so their heights were decreased after modification. That is the reason why some white space is present above the walls. In comparison, the bottom three examples did not have white space as these examples contain side walls, and the heights of the side walls were not supposed to be modified. In addition, because of the outward direction of the draft angle of the side walls, the modified designs exceeded the original bounding boxes in the horizontal direction. Other constraints like height, width, draft angle, and round corner were successfully and correctly added to different sub-designs.

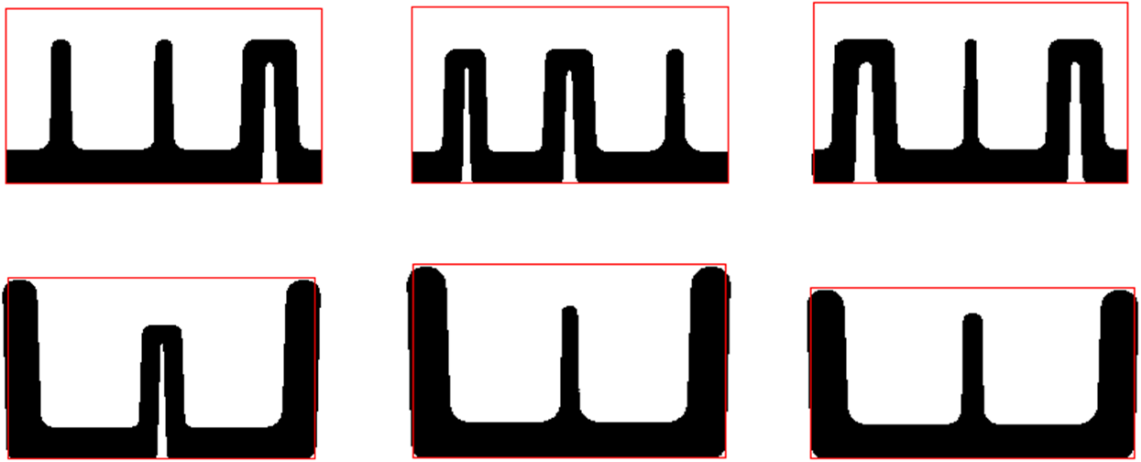

Figure 9. Examples of the results of McGAN

## 4.4. Comparison with other methods for unmanufacturable region modification

In this section, different image modification methods are tested to modify unmanufacturable regions, including PSPNet [16] and the Diffusion model [20]. The basic ideas of PSPNet and Diffusion models will be first introduced. Their performance is then compared with Pix2Pix.

The general goal of PSPNet is semantic segmentation, and for our task, it tries to separate the background and solid parts as shown in Figure 10a. One important feature of PSPNet is the utilization of dilated convolution, which is a modified version of convolution to include "holes". Suppose that the dilation rate of the dilated convolution is $D \times D$ and the kernel size is $K \times K$. Then, the dilated convolution will utilize $[D \times (K-1) + 1] \times [D \times (K-1) + 1]$ pixels as it will convolve every $D$ pixels while omitting the internal $(D-1)$ pixels. The benefit of dilated convolution lies in that it can have a larger receptive field with fewer parameters and it is efficient for implementation. More details of the dilated filter can be found in [55]. Another strength of PSPNet is attributed to the application of the "Pyramid Pooling Module" as shown in Figure 10b. After applying a CNN to extract the feature map, different average pooling sizes are utilized. For the feature map, it is firstly divided into $m \times m$ sub-regions where $m = 1,2,3,4$. Then, for each sub-region, average pooling, i.e., calculating the average value of the sub-region, is applied. After upsampling and concatenation with the original feature map, the pyramid pooling module is constructed. Pyramid Pooling can preserve more local information compared with Max-Pooling, and this local information will facilitate the next tasks of classification, segmentation, and modification.

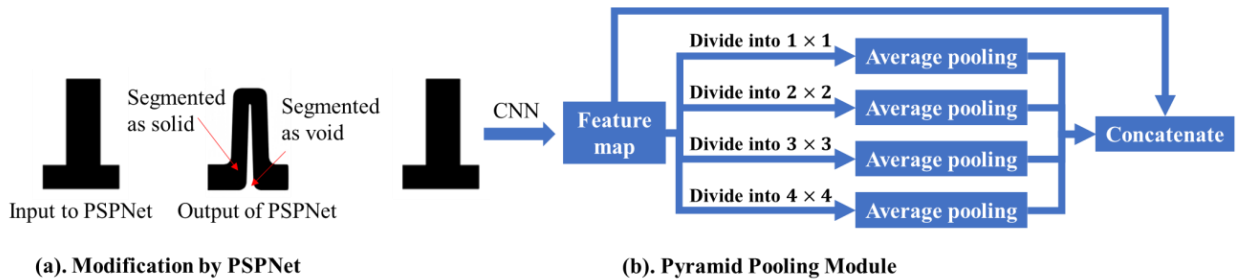

Figure 10. PSPNet for modifying unmanufacturable regions and Pyramid Pooling Module

The results of PSPNet can be found in Figure 11. The results show that PSPNet correctly modified the height and width of the middle wall, and it also tried to add a draft angle to the middle wall. However, it had some problems in rounding corners. It successfully added round corners at the intersection between the upper wall and the bottom wall, but failed to add round corners at the top region of the vertical wall. In addition, it seemed that PSPNet also tried to add draft angles at the left and right sides of the bottom wall, which is undesirable. In comparison, Pix2Pix did not have these problems and added the round corner to the correct position. As a result, it is evident that Pix2Pix performed better than PSPNet for modifying unmanufacturable regions.

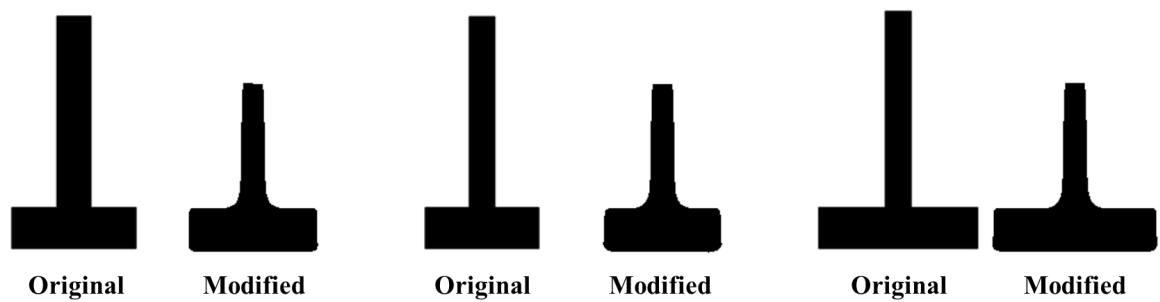
Figure 11. Unmanufacturable design modification by PSPNet

Another method utilized for comparison is the diffusion model, in particular, Denoising Diffusion Probabilistic Models (DDPM) [20]. DDPM modifies existing unmanufacturable designs by simulating the diffusion process. It is composed of two processes: forward process and reverse process as shown in Figure 12. In the forward process, Gaussian noise is gradually added to the manufacturable design until the final output becomes fully Gaussian noise. This process of adding noise may take approximately 1,000 iterations, and no training parameters are involved in the forward process. In the reverse process, the Gaussian noise generated by the forward process is concatenated with the unmanufacturable design. Then, a neural network with the UNet structure [15] is trained to remove the noise to generate the manufacturable designs. In the inference process after the training process, an image composed of random Gaussian noise is generated, and it will be concatenated with the unmanufacturable design to be modified. Then, following the reverse process, the noise will be removed until the final manufacturable design is generated.

Examples of DDPM in modifying unmanufacturable designs are shown in Figure 13. It seems that DDPM was less stable than Pix2Pix. To be more specific, when the upper wall was originally very thin, the modified upper wall disappeared as shown in the middle example. In addition, as shown in the right example, the round corner seemed to be a little asymmetric in the top regions. One other obvious drawback of DDPM lay in the slow inference time. Because the inverse process of DDPM iterated around 1,000 times, the inference speed of DDPM was 1000 times slower than Pix2Pix. More investigations may be required to further apply DDPM for modifying unmanufacturable designs.

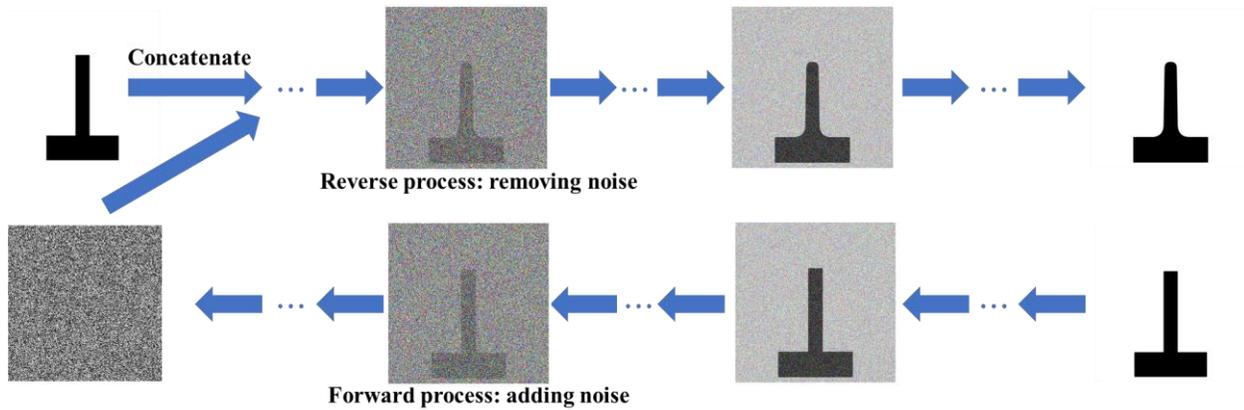

Figure 12. DDPM for modifying unmanufacturable designs

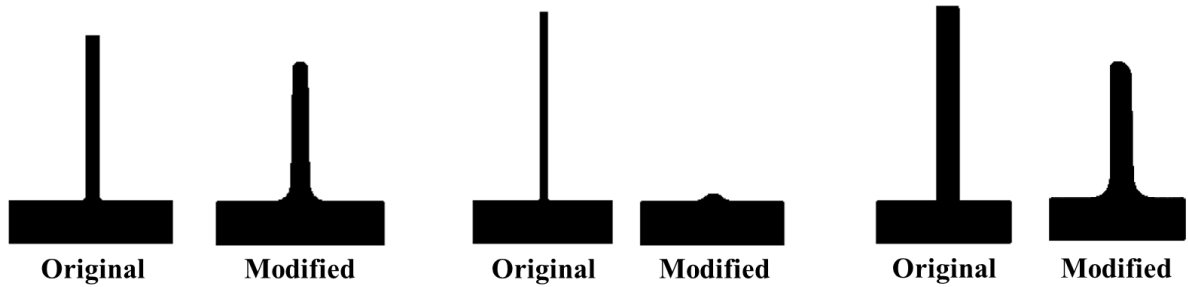

Figure 13. Unmanufacturable design modification by DDPM

## 4.5. Different "scales" for unit thickness

In the previous studies, the same scale was utilized for unit thickness. That is, the bottom wall has the same thickness in all training and testing examples. In the dataset for Mask R-CNN, the scale is $\frac{10 units}{256 pixels}$, while the scale is $\frac{6.6 units}{256 pixels}$ in the dataset for Pix2Pix. In this section, different scales are utilized in different examples, which means that unit length may take different numbers of pixels leading to different bottom wall thicknesses. As shown in Figure 14(a), the thickness of the bottom walls in the two examples are both taken as unit lengths, but the number of pixels for unit length in the left example is greater than that on the right. This is equivalent to different scales in CAD drawings. Based on these different scales, the rule for modifying the height of the upper wall is the following: the height of the upper wall should be shorter than or equal to four times the height of the bottom wall. When the Pix2Pix was trained with examples of different scales to modify height and width and was also tested on examples of different scales, some problems arise as shown in part (b) of Figure 14. As shown in the top example, the heights and thicknesses of the upper walls were modified. However, the left-most wall's height was not updated correctly; the top of the wall was not connected from the rest of the wall. Similar problems also arose for side walls as shown in the bottom example.

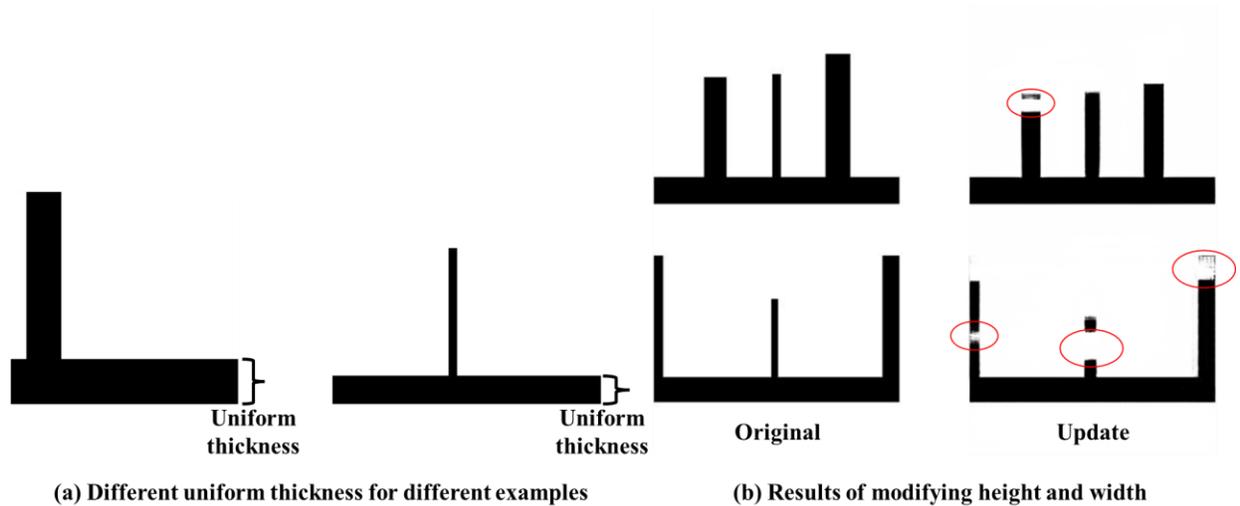

Figure 14. Modifying "height and width": training with a dataset composed of different scales for unit thickness

Engineers have the capability to recognize scales within designs and make design adjustments in accordance with those scales and manufacturing rules. In the most optimal scenario, we envision a well-trained Pix2Pix model being able to adeptly accommodate various scaling factors by emulating the expertise of experienced engineers. However, this problem seemed to be difficult for current state-of-art neural networks for image-to-image translation. This is an interesting future research direction to automatically modify unmanufacturable designs with different scales. In this paper, to avoid this problem, the same scale was utilized for uniform thickness.

### 4.6. Shorter segmentation mask

In previous experiments, the whole part design was segmented into regions with one thin, thick, or side wall. As a result, the bottom walls were generally relatively longer as shown in the top example compared with the bottom example of Figure 15. These long bottom wall segments may not be necessary as they will not impact the modification of unmanufacturable designs. To avoid long bottom walls, a short wall segmentation scheme was utilized as shown in the bottom example of Figure 15. Two obvious gaps are highlighted in the bottom example of Figure 15 by the two red bounding boxes. This is a better segmentation scheme from our perspective as only important regions that need to be modified are chosen and sent to Pix2Pix. In comparison, these long segments that do not require modification are left unchanged and no time is spent needlessly processing them. Based on the techniques of deleting repeating bounding boxes in section 4.1, all the repeating bounding boxes with $IOU \geq 0.2$ were deleted. For the tested 500 complex designs, all 1,500 single wall regions, with short bottom walls, were correctly and efficiently detected.

After all sub-designs were correctly identified, they were sent to Pix2Pix for modification for manufacturability improvement. To train Pix2Pixs, 12,000 new examples with short bottom walls (4,000 each for thin, thick, and side walls, respectively) were generated and utilized to train three separate Pix2Pixs for modifying the wall types. The results of the newly trained Pix2Pixs in modifying thin, thick, and side walls with short bottom walls can be found in Figure 16. All of them were modified correctly to facilitate manufacturing. The resizing and recombination scheme was applied to combine manufacturable sub-designs into the final manufacturable designs as shown in Figure 17.

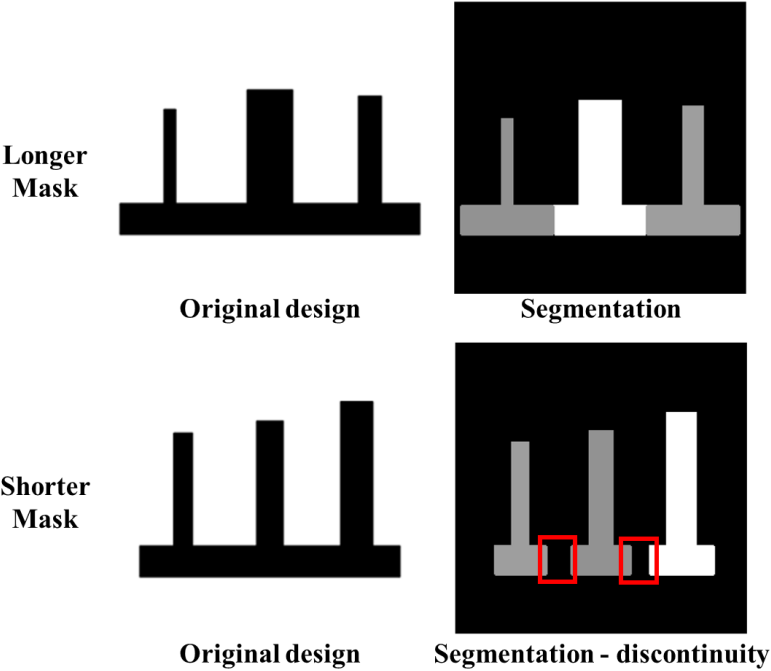

Figure 15. Two schemes of segmentation mask

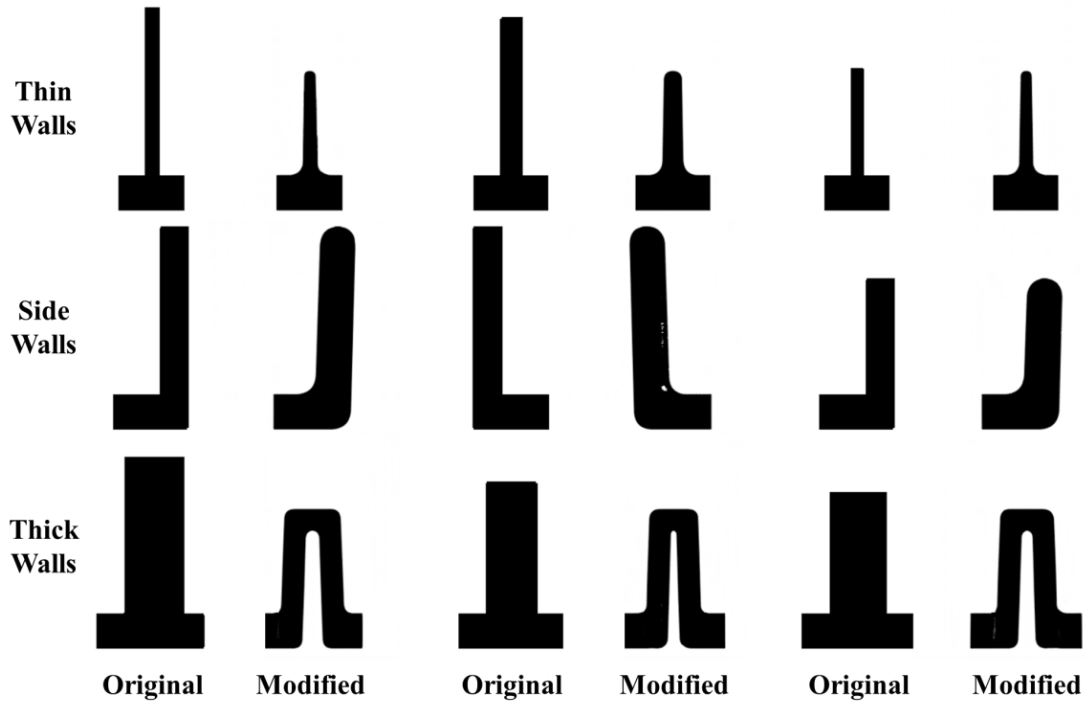

Figure 16. Pix2Pix modifies unmanufacturable sub-designs with short bottom walls

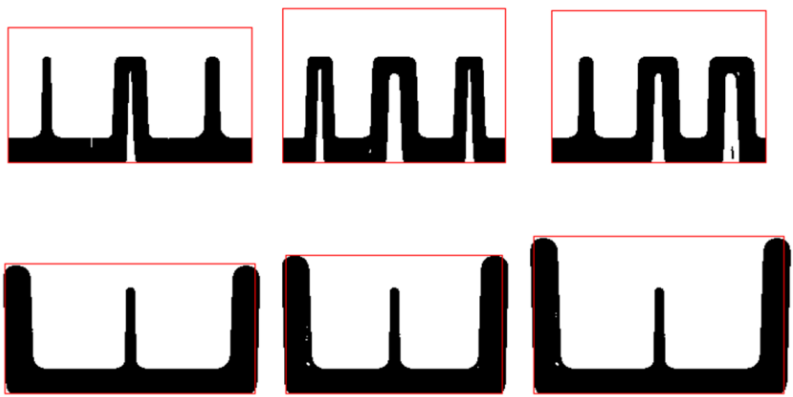

Figure 17. Results of McGAN on designs with short bottom walls for segmentation

### 4.7. More complex tests on five walls

In this section, more complex tests with five vertical walls are conducted to show the generalizability of McGAN. To begin with, the examples for training and testing Mask R-CNN are shown in Figure 18. On the left side, the original designs with five walls are shown, while their corresponding segmentation masks are shown on the right side. The longer segmentation mask was utilized, which is consistent with section 4.5 as the gap between two closely adjacent walls can be quite small. Five hundred examples with 5 walls each were used to test McGAN. Using the trained Mask R-CNN all 2500 vertical walls were successfully detected, segmented and classified. The derived walls were then sent to Pix2Pix for modification of height, width, draft angle and round corner. Examples of thin walls, side walls, and thick walls are shown in Figure 19. The "Original" are the inputs to Pix2Pix, while the "Modified" are the corresponding outputs of Pix2Pix after modifications. Eventually, the modified manufacturable walls were recombined and the complete manufacturable examples were derived and shown in Figure 20. The modified designs have proper heights, widths, draft angles and round corners. In addition, the red bounding boxes have dimensions of the original examples. When observing the top regions of the bounding boxes, some empty space exists that was caused by the decreased heights of thin and thick walls. Lastly, by comparison of the dimensions between inputs and outputs of McGAN, we can conclude that the heights and widths of original walls were properly modified, while other dimensions were properly maintained. This test strongly proves the generalizability of McGAN on more complex examples.

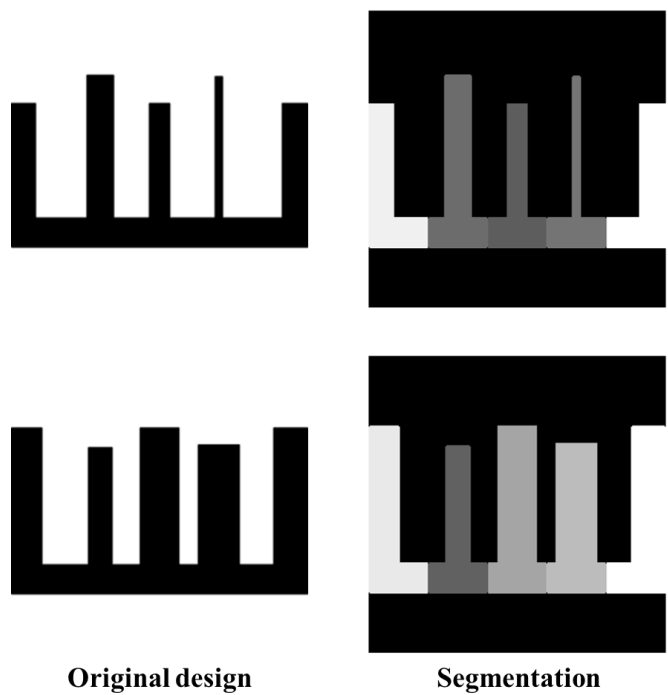

**Original design**  **Segmentation**

Figure 18. Original design (left) and segmentation (right) for more complex examples with five walls

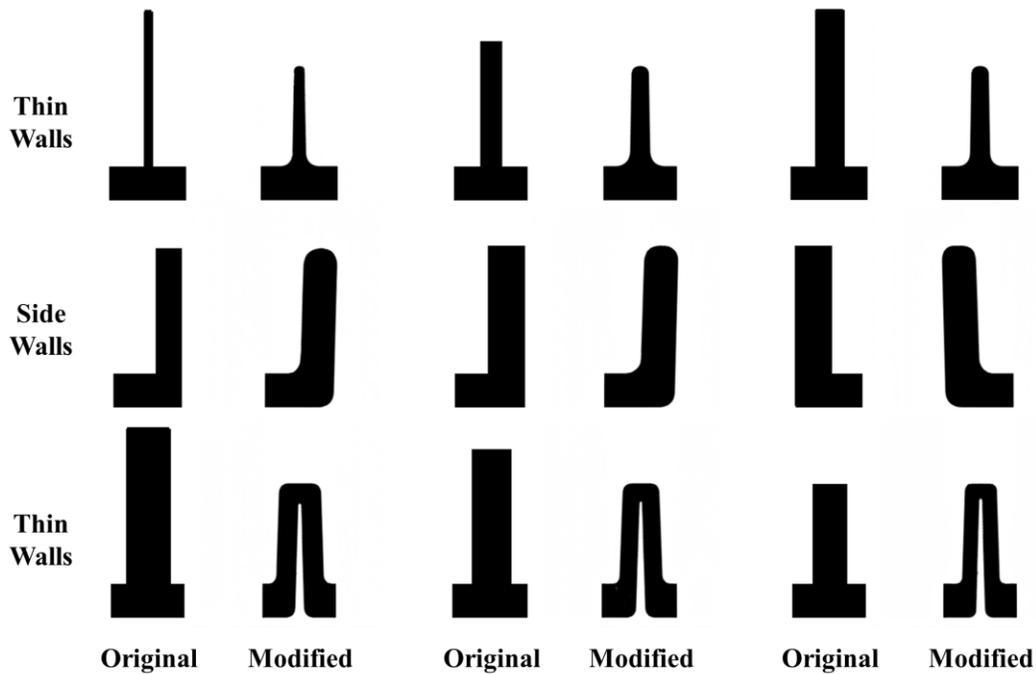

Figure 19. Pix2Pix for modifying segmented walls

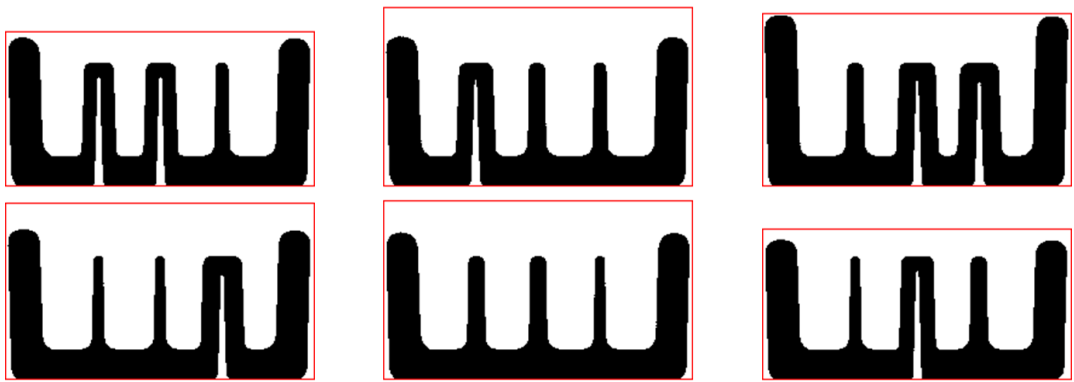

Figure 20. Recombination of modified walls to form manufacturable complex designs

### 4.8. Assessment of Partially Unmanufacturable Designs

In our previous experiments, we addressed modifications to entirely unmanufacturable designs. In the current study, depicted in Figure 21, our focus shifts to components that are partially unmanufacturable. These designs include regions where certain elements such as thin walls, thick walls, and side walls are manufacturable, whereas others remain unmanufacturable.

To address these challenges, we trained the Mask R-CNN model to identify both manufacturable and unmanufacturable features. During the training phase of the Pix2Pix model, we included both manufacturable and unmanufacturable walls as inputs, as illustrated in Figure 22. If the input corresponds to a manufacturable wall, the output is expected to replicate the input precisely. Conversely, if the input wall is deemed unmanufacturable, the Pix2Pix model aims to alter it to a manufacturable state. Throughout the testing phase, unmanufacturable walls were modified to meet manufacturability standards, while original manufacturable walls were preserved.

Subsequently, the outputs from the Pix2Pix model were reassembled into fully manufacturable structures, as demonstrated in Figure 23. Notably, in certain instances, the bounding boxes of the original designs exceeded those of the modified versions due to the presence of unmanufacturable tall thin or thick walls, which were subsequently adjusted in the Pix2Pix processing.

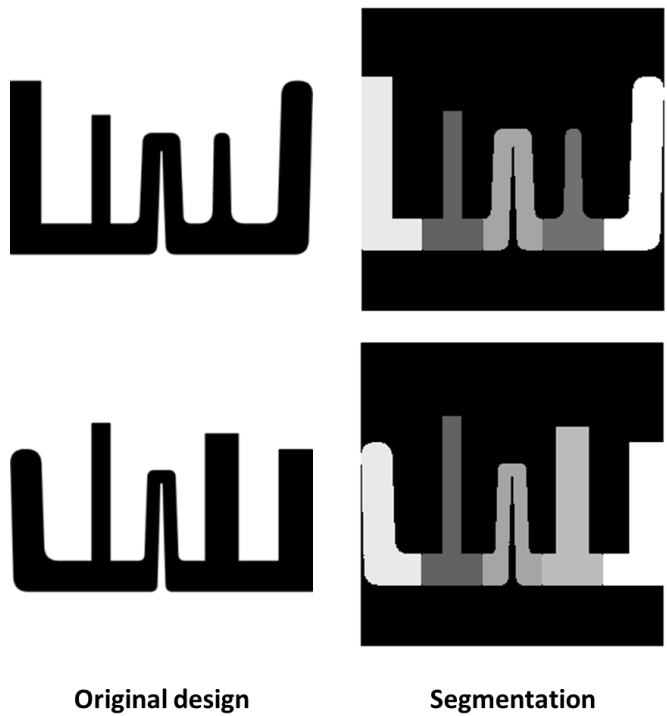

**Original design**      **Segmentation**

Figure 21. Original design (left) and segmentation (right) for partially unmanufacturable examples with five walls

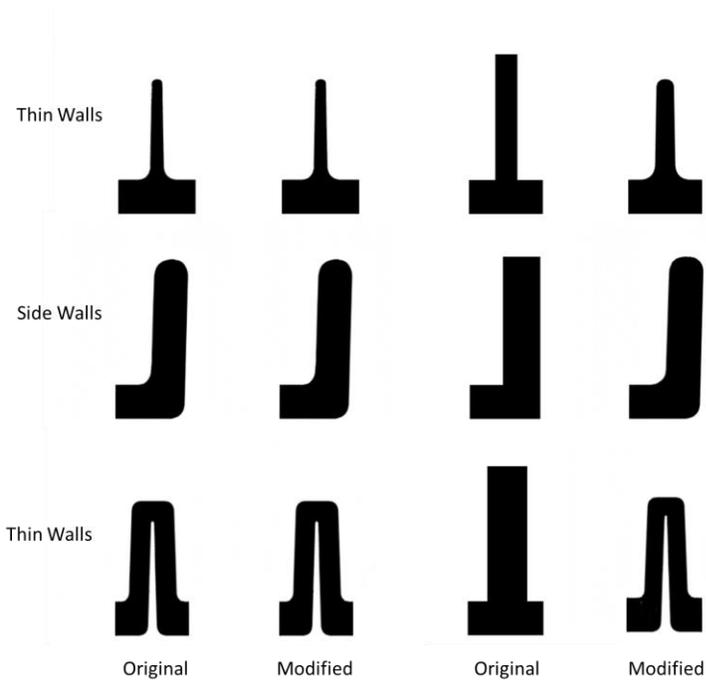

Figure 22. Pix2Pix for modifying segmented walls (both manufacturable and unmanufacturable)

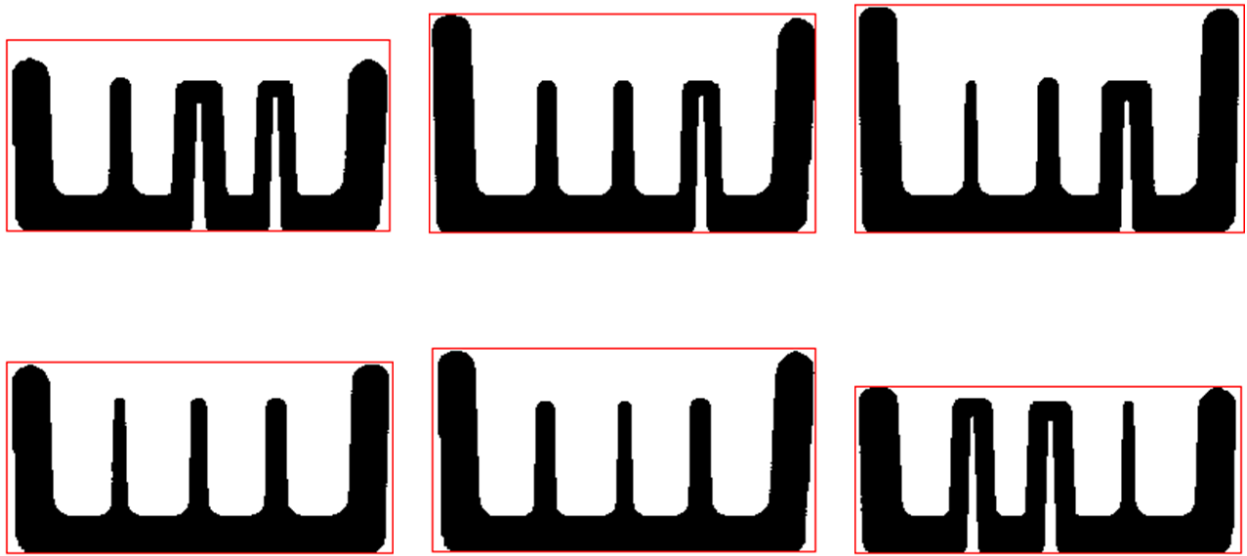

Figure 23. Recombination of modified walls to form manufacturable designs

### 4.9. Computation cost analysis

In this section, the training and testing times of McGAN will be investigated. All these experiments were conducted with an NVIDIA GeForce RTX 3070, with 8GB memory in total. The memory was enough for the training and testing of McGAN. Because the reintegration module takes nearly no time, we will only pay attention to the training and testing time of Mask R-CNN and Pix2Pix. The computation costs for Mask R-CNN and Pix2Pix can be found in Table 3.

The training for Mask R-CNN was conducted on 5,000 examples and 10 epochs were included. The total time for training Mask R-CNN was 100 minutes, which is equivalent to 10 minutes per epoch. In comparison, the test inference time for Mask R-CNN was 0.79 seconds per design.

Three different Pix2Pixs were trained for thin, thick, and side walls respectively. The number of training examples for each Pix2Pix was 4,000. The total number of training epochs was 200. The total training time was 500 minutes, which is equivalent to 2.5 minutes per epoch. The test inference time for Pix2Pix was 0.026 seconds per design.

Table 3. Computation cost analysis of Mask R-CNN and Pix2Pix

| Model | Training examples | Epochs | Training time (minutes) | Training time per epoch (minutes) | Testing time per design (seconds) |
|---|---|---|---|---|---|
| Mask R-CNN | 5000 | 10 | 100 | 10 | 0.79 |
| Pix2Pix | 4000 | 200 | 500 | 2.5 | 0.026 |

## 4.10. Discussion

Results show that McGAN modified the unmanufacturable designs and generated their corresponding manufacturable counterparts. In this paper, seven different experiments were performed on McGAN. The first experiment utilized Mask R-CNN to apply instance segmentation on complex wall designs. Instance segmentation includes three tasks: object detection, object classification, and segmentation. The results of object detection and object classification are utilized in subsequent tasks. The second task trained three separate Pix2Pixs to modify unmanufacturable regions of thin walls, thick walls, and side walls respectively. We also trained one Pix2Pix to modify unmanufacturable regions of all different types of walls together. However, the test results indicated that training three Pix2Pixs to modify each type of wall separately worked better. In the third task, the framework of McGAN was constructed with Mask R-CNN for instance segmentation, resizing to expand the sub-designs, Pix2Pix to modify unmanufacturable sub-regions, resizing to shrink the sub-designs and recombination to construct the final manufacturable designs. This framework modified complex designs to make them manufacturable, and it was further extended to even more complex parts. In the fourth experiment, a comparison between Pix2Pix and two other methods for design modification, PSPNet and DDPM, was conducted. From the results, we concluded that Pix2Pix performed best from the perspective of the modified design quality and robustness. In the fifth experiment, different scales were utilized for unit thickness across different examples. Though consistency across different scales was a desired property for Pix2Pix, our results showed that it struggled with different scales for unit length. In the next experiment, a shorter bottom wall length was utilized for training Mask R-CNN and Pix2Pix, and the modified designs were also appropriate for manufacturing. The seventh test worked on more complex designs with five walls, and the success of this experiment showed the strong generality of McGAN to more complex designs. The last experiment tested designs that were partially unmanufacturable. The unmanufacturable regions were modified and the original manufacturable regions were maintained. As a result, the designs were selectively altered to ensure manufacturability, as desired.

From the results of the first four experiments, McGAN, composed of Mask R-CNN, Pix2Pix and reintegration, succeeded in modifying existing unmanufacturable designs to make them manufacturable and Pix2Pix performed better than other methods from the literature from the viewpoint of manufacturability improvement. However, the fifth experiment showed that more research is required on how to accommodate different scales for different designs. If this problem can be solved, the application field of McGAN will be greatly expanded.

Current work focuses on injection molding examples, with attention to parts composed of different types of simple walls. However, there are several limitations of this work. The paper only considers the cases of different walls in housing parts, which are manufactured through injection molding. In comparison, other types of parts manufactured by injection molding with different manufacturing features like plastic bottles are ignored. In addition, other manufacturing processes are omitted like laser-cut and machining. But the idea of McGAN can be extended easily to more complex designs and rules. Importantly, the three injection molding rules implemented using Pix2Pix are representative of many DFM rules. Specifically, the three rules 1) add a geometric feature (rounded corner), 2) modify feature shape (draft angle), and 3) adjust feature dimensions (aspect ratio). Variations of these rules should ensure that McGAN applies to most other manufacturing processes. However, we recognize that more powerful instance segmentation tools may be required, particularly for more complex designs.

Another interesting research direction will be the combination of DFM using McGAN with other generative design approaches focused on achieving desired functionality, to ensure that designs are both functional and manufacturable. In this work, the functionality of designs after modification is not considered. It is not clear if the McGAN approach can be extended to include functionality, or if design for functionality approaches could be extended to incorporate McGAN technologies. Alternatively, it may be necessary to iterate between separate design for functionality and DFM approaches.

Another possible research direction lies in dealing with feature coupling. Suppose there are two features that are identified. The two features can be modified locally, while they may have some spatial relationship that restricts the manufacturing process globally. For instance, machining tool accessibility at a specific point will become problematic when adjacent regions undergo modifications. It is worth noting that this paper does not address this particular issue, leaving room for further exploration and future research.

Current work focuses on McGAN in the 2D domain. Since mechanical designs are 3D, it is necessary to extend McGAN into the 3D domain. Initial experiments indicate that this extension is straightforward, although significantly more computationally expensive. Voxel patterns of manufacturing rules will be more complicated than the simple 2D rules that have been utilized in this work so far, which may raise challenging research issues. To tackle the computational challenges of 3D McGAN, there are two promising solutions. The first approach involves modifying designs at a lower resolution and employing an additional super-resolution neural network to enhance the resolution of the modified designs. The second avenue worth exploring is leveraging sparse convolution, a technique that holds the potential to significantly cut down on computation costs.

As a final comment, this work represents an initial study of generative design for manufacturability. Although promising, significant work remains to improve and generalize the methods. Some research issues were highlighted that should guide future investigations.

## 5. Conclusion

To facilitate the manufacturing process, automatic generative design for manufacturability should be established. Previous work focused on automatic generative design for functionality, while the research on design for manufacturability was scarce. In this work, a novel conditional GAN framework, called McGAN, was constructed to transform unmanufacturable designs into manufacturable designs. McGAN is mainly composed of three parts: Mask R-CNN, Pix2Pix, and reintegration. Mask R-CNN is utilized to detect, classify, and segment a complex design into multiple simple sub-designs. Given each sub-design, Pix2Pix is applied for modification; i.e., Pix2Pix takes the original unmanufacturable design as input and outputs its corresponding manufacturable design. Lastly, the reintegration of multiple modified manufacturable features forms the final modified manufacturable designs. To show the strength of McGAN, different experiments were conducted to test Mask R-CNN on detecting, classifying, and segmenting complex designs. In addition, we tested Pix2Pix on modifying different types of unmanufacturable designs consisting of combinations of thin, thick, and side walls. From there, the complete McGAN approach was tested on modifying complex designs to make them manufacturable as a whole. A comparison of Pix2Pix with other manufacturability modification methods including PSPNet and DDPM demonstrated the superior performance of McGAN. Tests on different scales for design features faced some challenges and identified some future research directions. Experiments on short segmentation masks showed the possibility of automatically modifying more complex designs by extracting only important information for modification. The test on more complex designs with five vertical walls showed the strong generality of McGAN.

Based on this research, the following conclusions can be drawn:

- Deep neural networks represent a promising technique for automatically modifying unmanufacturable designs and generating their corresponding manufacturable designs. More specifically, the combination of Mask R-CNN and Pix2Pix technologies, along with reintegration, to form the McGAN approach can automatically modify complex designs to improve their manufacturability.
- Mask R-CNN, or more generally instance segmentation, can automatically simplify the DFM task by decomposing a complex design into a set of simple sub-designs.
- Pix2Pix, for image-to-image translation, can automatically modify unmanufacturable features of original designs.

The results of McGAN are promising as modified designs are of high quality from the viewpoint of manufacturability and this framework should be extendable to other manufacturing processes.

Future research should focus on several other directions. The first direction lies in how to solve the problem of different scales of unit length, described in section 4.5. Extension of the McGAN framework to other designs and manufacturing processes is of great interest. In addition, the combination of design for functionality and manufacturability should receive more attention as real part designs need to satisfy both. It is likely that significant research issues will arise on this topic. Extending McGAN to the 3D domain is another needed direction, but faces computation cost challenges. These important factors will be considered in our future research work.

**Acknowledgment**

The authors gratefully acknowledge support from the National Science Foundation through EAGER grant #2113672, and the Future Manufacturing Research Grant #2229260. The author also thanks Dr. Junbo Peng's help in modeling DDPM. Any opinions, findings, and conclusions or recommendations expressed in this publication are those of the authors and do not necessarily reflect the views of the National Science Foundation. To the best of the authors' knowledge, there are no conflicts of interest. The data and code will be shared on GitHub once the paper is accepted.